\documentclass[times,twocolumn]{article}
\usepackage[utf8]{inputenc}
\usepackage{amsmath}
\usepackage[margin=0.5in]{geometry}
\usepackage{graphicx}
\usepackage[english]{babel,varioref}
\usepackage[font=small, labelfont=bf]{caption}
\usepackage{subcaption}
\usepackage{multimedia}
\usepackage{import}
\usepackage{multicol}
\usepackage{amsfonts}
\usepackage{listings}
\usepackage{color}
\usepackage[
  maxcitenames=2,
  doi=false,
  backend=biber,
  firstinits=true,
  style=numeric-comp,
  sorting=none,]
{biblatex}
\usepackage{tikz}
\usetikzlibrary{external,positioning} 
\usepackage{siunitx}

\author{
Erik Wallin\textsuperscript{1}\thanks{Corresponding author: erik.wallin@umu.se},
Viktor Wiberg\textsuperscript{1},
Martin Servin\textsuperscript{1}\\
\textsuperscript{1}Department of Physics, Umeå University
}

\title{Multi-log grasping using reinforcement learning and virtual visual servoing}

\begin{document}
\maketitle

\begin{abstract}
We explore multi-log grasping using reinforcement learning and virtual visual servoing for automated forwarding in a simulated environment.
Automation of forest processes is a major challenge, and many techniques regarding robot control pose different challenges due to the unstructured and harsh outdoor environment.
Grasping multiple logs involves various problems of dynamics and path planning, where understanding the interaction between the grapple, logs, terrain, and obstacles requires visual information.
To address these challenges, we separate image segmentation from crane control and utilise a virtual camera to provide an image stream from reconstructed 3D data.
We use Cartesian control to simplify domain transfer to real-world applications.
Since log piles are static, visual servoing using a 3D reconstruction of the pile and its surroundings is equivalent to using real camera data until the point of grasping.
This relaxes the limits on computational resources and time for the challenge of image segmentation and allows for collecting data in situations where the log piles are not occluded.
The disadvantage is the lack of information during grasping.
We demonstrate that this problem is manageable and present an agent that is 95\% successful in picking one or several logs from challenging piles of 2--5 logs.
\end{abstract}

\providecommand{\keywords}[1]
{
  \small
  \textbf{\textit{Keywords---}} #1
}

\section{Introduction}
Automatic loading of multiple logs requires visuomotor control of a crane manipulator in a complex environment.
This involves challenges in collecting and interpreting visual information for grasping and crane motion planning to handle obstacles, grapple-pile dynamics, and external conditions.
Improvements in efficiency and automation are important for the forestry industry's role in sustainability, but these pose major challenges due to the unstructured and harsh outdoor environment.
The rough terrain with various obstacles, shaking, wear and tear of equipment, and exposure to light, weather, and seasonal conditions pose different challenges compared to the environment of conventional robot control, in particular for vision-based systems.
A forwarder spends most of its time picking up logs~\cite{lundback2022economic}, and it is crucial for high efficiency to lift multiple logs with each grasp without exceeding the maximum lift capacity of the crane.
This requires detailed and unobstructed information about the piles and the environment and makes data collection, segmentation, and crane control significant challenges that must be addressed to enable reliable and robust autonomous forwarding.

Driven by the global trend of big data and the progress in machine learning, the forestry industry is experiencing an increase in the collection and availability of large amounts of data.
Harvest areas can be scanned from the air and the ground, and ground and trees can be segmented~\cite{Axelsson1999,elmqvist2001}, allowing for detailed terrain maps for path planning~\cite{wallin2022learning}, for example.
Harvesters are increasingly being equipped with high-precision positioning systems and can store geospatial information of the felled logs~\cite{lindroos2019advances}, as well as the paths travelled.
This opens up possibilities for autonomous forwarding and increased efficiency in forestry.
Removing the operator from the vehicle also relaxes the economy, ergonomic, and design constraints.
While fully autonomous forwarding is a challenge, more imminent scenarios include operator assistance, remote-controlled machines, or partially autonomous functions.

The process of grasping logs in forestry is related to the general field of robotic grasping, which has been extensively explored in recent years~\cite{caldera2018review,levine2018learning,kleeberger2020survey}.
However, there are differences that make log grasping a special case, most notably regarding grasping multiple objects, the unstructured forest environment, the electro-hydraulic crane actuation, the system size, and exposure to the elements.
For the specific application of log grasping and autonomous forwarding, there are good solutions for crane motion planning and control~\cite{ortiz2014increasing, taheri2022nonlinear}, not considering grapple-log interaction or surrounding obstacles.
Reinforcement learning (RL) control has also been proven to be effective for the same task in simulations, grasping a single log with known pose~\cite{andersson2021reinforcement}.
Transferring such joint-level RL control to a real system is, however, a problem due to the simulation bias when the electro-hydraulic circuit~\cite{zhao2020sim2real, wiberg2023sim} has not been precisely modelled.
\textcite{dhakate2022autonomous} shows how joints can be modelled and dynamics learned using RL to enable Cartesian control.
Actuator dynamics are specific to each machine, non-intuitive for humans, and difficult to interface with other control systems or human operators for shared control of crane operation~\cite{hansson2010semi}.
Cartesian control, on the other hand, can be seen as a common interface, which is more intuitive and more easily interfaced with other systems.
Considering the grapple, logs and obstacles, there is a need for visual input to take their configurations and interactions into account.
Logs may be partially overlapped or interlocked, and successful grasps may depend on small geometry details that affect the interaction between the grapple and the logs.
At the same time, the terrain and obstacles, such as trees and rocks, make the grasping task more than a grasp-pose estimation problem, involving a crane control problem with grasp dynamics and path planning.
There are methods for log detection~\cite{ainetter2021depth,fortin2022instance}, but varying conditions and occlusion make real-time segmentation difficult and hinder continuous crane and grasping control.
There are, however, promising experiments where segmentation has been used to identify grasp poses.
\textcite{hera2023exploring} shows sparks of early autonomous forwarding in practice, picking single logs along a path on flat ground in concept machine experiments.
\textcite{ayoub2023grasp} develop a grasp planning algorithm which is successfully tested on a physical crane to grasp single or multiple logs on flat ground.
Logs are segmented and modelled in a simulator to produce depth-camera images, from which a grasp pose is generated by a convolutional neural network (CNN).

\begin{figure}[ht]
\centering
  \includegraphics[width=0.48\textwidth]{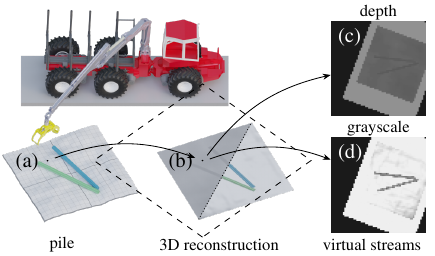}
  \caption{\label{fig:virtual_camera} Illustration of the virtual camera setup, showing (\textbf{a}) the actual pile, (\textbf{b}) the corresponding 3D reconstruction, and (\textbf{c},\textbf{d}) the depth- and greyscale virtual streams.
  The position of the virtual camera is represented by a dot, with the orientation and extent illustrated by the dashed square.}
\end{figure}

Visual information for continuous crane and grasping control should provide a good overview and be unobstructed, including occlusion by the crane and grapple.
It would also be beneficial to collect visual data during moments with good visibility or to combine data from different times and perspectives.
Another option would be to separate segmentation and control, using specialised systems for each.
Considering this, we define a \emph{virtual camera} as a sensor that generates a stream of 2D data originating from a 3D reconstruction; see Fig.~\ref{fig:virtual_camera}.

To address the challenge of collecting and using visual data for control in challenging forest environments, we explore using reinforcement learning and virtual visual servoing for multi-log grasping.
We utilise Cartesian control to simplify the typical reinforcement learning problems of simulation-to-reality (sim-to-real) transfer and interfacing with other control systems or human operators.
To address the issue of occlusion in visual servoing for crane control, we utilise a virtual camera, where the underlying 3D reconstruction data can be captured where there is no obstruction.
This also enables combining data from different times or perspectives and removes the need for real-time segmentation, allowing more time and computational resources for this task.
We train agents using multibody dynamics with frictional contacts, with a reward signal designed to give dense feedback from the camera data.
In addition, we investigate ways to gain insights into learned behaviours with a focus on the use of image data.

\section{Method}
To test control from 3D reconstructed data using virtual cameras, we train an agent to grasp multiple logs using model-free RL.
Application in practice would require segmenting logs and removing disturbing background from real image data~\cite{ainetter2021depth,fortin2022instance}.
Here, we work with piles generated to match such corresponding output.
We generate log piles and simulate a forwarder using multibody dynamics with frictional contacts using the physics engine AGX Dynamics~\cite{AGX2023}.

\subsection{Piles and Virtual Camera}
\begin{figure}[ht]
\centering
  \includegraphics[width=0.50\textwidth]{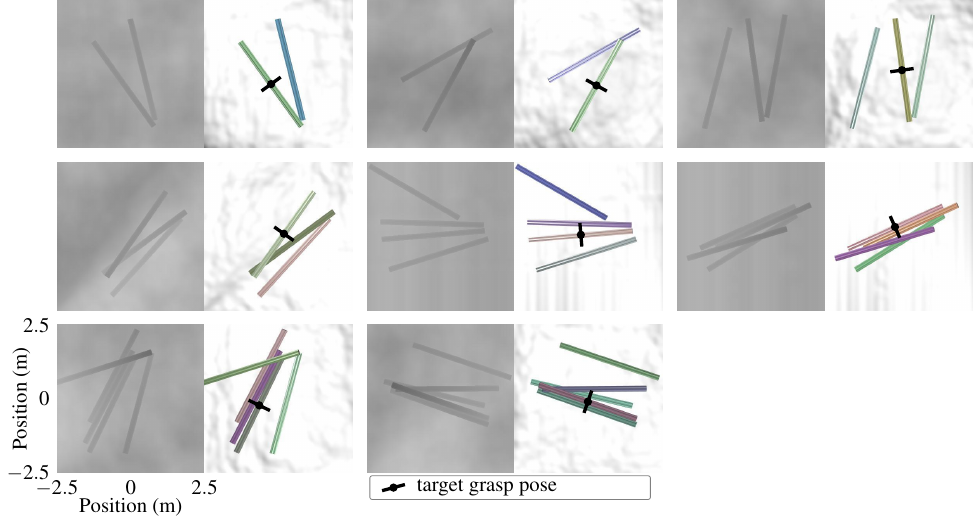}
  \caption{\label{fig:pile_example} Example of piles, with corresponding depth and RGB images for eight piles with 2--5 logs. The elevation difference of the used terrains ranges from 0.2~m to 0.8~m, with a mean of 0.4~m.}
\end{figure}

We generate uneven terrains using Perlin noise~\cite{perlin1985image}, as $5 \times 5$~m$^2$ patches, and form disordered piles with 2--5 logs by stacking logs vertically with random displacements and rotations in the horizontal plane, and letting them fall to the ground.
To emulate output from log segmentation, the ground is coloured in a uniform bright colour, and colour and depth (RGB-D) images are generated using an orthographic camera placed straight above the pile, as seen in Fig.~\ref{fig:pile_example}.
The displacement components and rotation for the logs were sampled from Gaussian distributions centred around zero with $\sigma_{\text{pos}}=0.5$~m and $\sigma_{\text{rot}}=0.25$~rad, determined empirically to achieve varying and challenging piles.
To make logs less prone to rolling, they are modelled by two overlapping square cuboids with a relative rotation of $45^{\circ}$.
We delimit ourselves to fixed-sized and shaped logs, using cuboids that are 3.5~m long and $\sqrt{2}/10$~m thick to emulate logs with a diameter of 0.2~m and a mass of 112~kg.
Cases where logs do not relax quickly, were discarded by comparing the mean log speed to a small threshold $\epsilon_v=5 \times 10^{-3}$m/s within 10~s.
The target grasp pose is set according to the position and orientation of the log closest to the combined log centre of mass position, which is not occluded by any other log; see Fig.~\ref{fig:pile_example}.

The aim of the virtual camera is to imitate the output of a real camera as if mounted on the grapple, but using segmented 3D reconstructed data; see Fig.~\ref{fig:virtual_camera}.
The relative position $\textbf{r}_{\text{rel}}$ and orientation $\phi_{\text{rel}}$ of the pile and the virtual camera are used to transform the RGB-D data to a virtual camera output stream.
To reduce the dimensionality of the camera data, the RGB data is converted to greyscale.
The RGB colours of the logs were sampled from small ($\sigma=10\%$) Gaussian variations around grey.
This makes all logs similar in greyscale, emphasising that logs must not be individually segmented.

The orthographic camera lacks perspective and is simply specified by its resolution and physical size.
We set the resolution to $64 \times 64$ pixels, and to mimic a field-of-view, we vary the camera size depending on the z-component of $\textbf{r}_{\text{rel}}$.
This is done by defining the camera sizes $s_{\text{far}}$ and $s_{\text{near}}$ at some \emph{far} (5~m) and \emph{near} (0~m) distances, and using linear interpolation in between.
The virtual camera is not limited to obeying the constraints of physics, as a real camera is.
This flexibility allows for exploring scenarios that may be challenging or unattainable to replicate in the physical world.
We explore $s_{\text{far}}=15$~m and $s_{\text{near}}=3$~m to keep an overview even at the grasp moment, when the grapple is close to the pile.
The RGB sensor data is independent of the distance to the pile, but the depth sensor data is rescaled to match the output of a real depth camera.
A major difference to a physical camera is that the underlying data of the virtual camera will not update after grasping.
We investigate how important the different observables are to the agent's behaviour at different stages of the grasping cycle.

\subsection{Crane Control and Calibration}

\begin{figure}[ht]
\centering
  \includegraphics[width=0.50\textwidth]{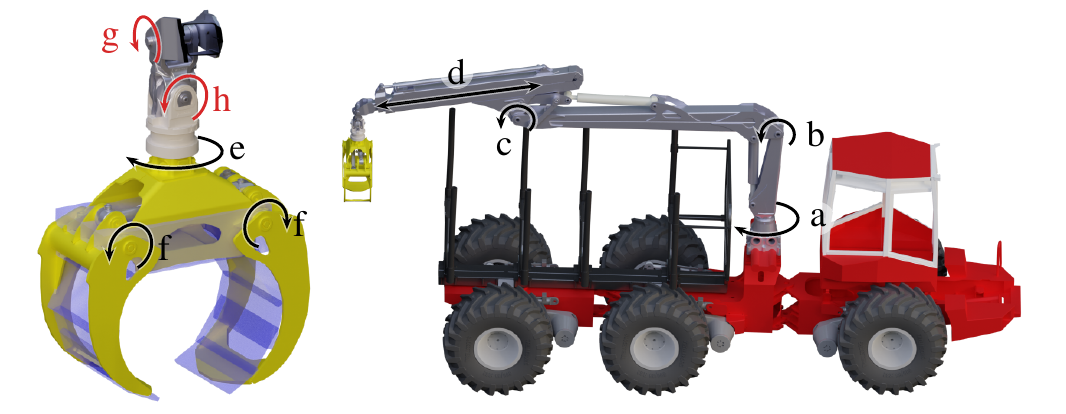}
  \caption{\label{fig:xt28} The \emph{Xt28} concept forwarder with the \emph{Cranab FC12} crane mounted.
    The semi-transparent blue boxes show the simplified grapple geometry.}
\end{figure}

The crane is a \emph{Cranab FC12} (Cranab AB) mounted on the \emph{Xt28} (eXtractor AB) pendulum arms concept forwarder; see Fig.~\ref{fig:xt28}.
It consists of 21 bodies and 26 joints, of which 6 are actuated.
The \emph{pillar} is connected to the \emph{base} and can rotate by an actuated hinge (a).
From the pillar, the \emph{main boom} is connected with a hinge (b) and a piston providing hydraulic power.
The \emph{outer boom} similarly connects (c) from the main boom, and the \emph{telescope} can extend (d) from the outer boom, powered by a piston.
The end-effector consists of a \emph{rotator} and a \emph{grapple}.
The rotator has one actuated hinge (e) for rotating the grapple and two hinges (g--h) that allow the grapple to swing.
The grapple opens and closes (f), powered by a piston.
To speed up simulations, the mesh geometry of the grapple is replaced by a similar but simplified geometry made up of 9 boxes, while the original geometry is kept for visuals.
Joint range and force limits are calibrated using data from the manufacturer~\cite{cranab-brochure} but are not experimentally confirmed.
Joint range limits are set using the maximum reach of the crane and illustrations/images of different configurations, and force limits are set guided by data of the lift capacity at some discrete crane configurations.
The lowest lift capacity, at the 8~m full extension, is 9.7~kN.
As the logs weigh 112~kg, the lift capacity is enough to lift 5 logs even at full extension easily.
To model the friction in the rotator hinges, we use weak lock constraints and tune the force limits and compliance until the damping of the swinging of the grapple appears physical and agreeable with video material.
The crane weighs 1630~kg, while the rotator and grapple weigh 249~kg together.

We implement \emph{Cartesian control}, where from a desired crane-tip velocity $\textbf{v}_{\text{crane-tip}}$ in Cartesian world frame coordinates, the corresponding target velocity of each joint is calculated with inverse kinematics~\cite{spong2008robot}.
As an alternative to joint-level control, Cartesian control is becoming increasingly common in commercial forest machines~\cite{john-deere-patent}.
Actuator dynamics are specific to each machine design, whereas Cartesian control could be seen as a layer of abstraction, exposing a common interface.
This increases the generality and simplifies implementation and sim-to-real transfer, removing the need for precise modelling of the electro-hydraulic crane actuation.
It also simplifies combining the control with human operators or other control systems~\cite{hansson2010semi}, e.g., for avoiding obstacles.

The Cartesian control problem for the described crane, with 4 degrees of freedom to control the 3 components of the crane-tip velocity, is an \emph{under-determined} system.
Thus, there is no inverse to the Jacobian describing how the crane-tip velocity is affected by the velocity of each joint, given some crane configuration; i.e., there can be (infinitely) many joint velocity solutions for a single crane-tip velocity.
This issue is addressed by defining a \emph{pseudo-inverse}, with weights for prioritising motion in different joints.
We define these as functions of the articulation of each joint, which are approximately constant but decrease to 10\% near the range limits.
This makes the system solvable, with solutions mostly within the physical limits of the actuators.

To simplify the modelling and not slow down simulations, we model the crane hydraulics using kinematic constraints instead of hydraulic- and electric circuit simulations.
For each actuator, the force/torque is determined as a solution of the multibody dynamics equation, considering the given limits on joint ranges and motor force.
To mimic the relatively slow motion of the hydraulics, the requested joint velocities are restricted by clipping in the range $[-1, 1]$~m/s (rad/s).

\subsection{Reinforcement Learning Control}
Reinforcement learning is a machine learning method in which an agent learns through trial and error.
It has proven successful on complex control problems with high-dimensional observations such as visual data, where otherwise conventional control systems have struggled.
The agent selects an \emph{action} based on a \emph{state} and its \emph{observation} of it.
A \emph{reward} signal is used to guide the learning towards desired state-action mappings~\cite{sutton2018reinforcement}.
RL has led to many impressive results, especially in games~\cite{mnih2015human}, but has yet to be widely used in real-world applications.
Compared to classical control methods, its main strengths are in complex planning tasks with long horizons and many degrees of freedom.

\subsubsection{Observation and Action}
The observation space consists of the virtual camera output and 16 scalar values concerning the crane, grapple, and target configurations.
The camera data is 64~$\times$~64 pixels with two channels.
To keep the idea of Cartesian control as a high-level interface, we chose not to include joint observations of the crane, i.e. angles/speed of the joints (a--d) in Fig.~\ref{fig:xt28}.
Instead, we provide the grapple's relative position, velocity, and speed with respect to the target.
However, details regarding the rotator and grapple are provided, with angles and angular speed for the rotation, swing (2 directions), and grapple opening.
Furthermore, to compensate for the lack of joint observations and not deprive the agent of all haptic sense, we provide a virtual load cell in the rotator.
This measures the grapple-load weight, normalised by subtracting and dividing with the empty grapple weight.
In practice, the crane configuration and the pressure in the hydraulic cylinder of the main boom could provide such force estimates.
Angle and speed observations for the grapple and rotator joints are scaled to $[-1, 1]$ using their respective limits, while other observations are clipped to $[-10, 10]$ to encompass the full range of the typical relative grapple position components.
The relative rotation of the grapple to the target angle has \emph{not} been included as one of the observations.
The motivation is to create a dense dependence on the camera data, where information on the angles of all logs compared to the grapple is contained.
We suggest that this increases the ability of the agent to analyse the camera data, which simplifies the learning process.

The action consists of five scalar values, where three represent the velocity components of the desired crane-tip velocity, and the other two represent rotating and opening/closing the grapple.

\subsubsection{Reward}
We have designed a reward function
\begin{equation}
  \label{}
  r = r_{\text{target}} + r_{\text{guide}} + r_{\text{energy}}
\end{equation}
that combines a sparse term related to overall success or failure with dense terms to aid learning from image data.
The sparse term $r_{\text{target}}$ is designed to become the dominant term, with the others to aid learning without too much biasing the final behaviour.
The relative contributions to the accumulated reward depend on the learned behaviour and cannot be immediately inferred.
For the trained agent, they are $92\%$, $10\%$, and $-2\%$, respectively.

We use zero-centred Gaussian functions for scaling, denoting these $G(x; \sigma) = e^{-0.5(x/\sigma)^2}$ for some measure $x$, or $G_{\sigma}$ for short.
The first term, $r_{\text{target}}$, is awarded only when the agent has achieved the target objective of grasping one or several logs and lifting them a sufficient height off the ground as
\begin{equation}
  \label{eq:reward}
  r_{\text{target}} = 25 G(x_{\Delta\text{grasp}}; \sigma_{\Delta\text{grasp}}) + 1.12 N_{\text{logs}}
\end{equation}
where $x_{\Delta\text{grasp}}$ is the proximity of the grapple to the centre of mass of the logs in the grapple, $\sigma_{\Delta\text{grasp}}=0.5$~m, and $N_{\text{logs}}$ is the number of logs in the grapple.

The second term in Eq.~(\ref{eq:reward}), $r_{\text{guide}}$, is a dense reward designed
to help the agent consistently learn to grasp logs, as
\begin{equation}
  \label{}
r_{\text{guide}} = r_{\text{stage}} G_{\Delta\text{tilt}} / N_{\text{steps}}
\end{equation}
where $G_{\Delta\text{tilt}}$ scales with the vertical tilt of the grapple, $\sigma_{\Delta\text{tilt}}=0.2$, $N_{\text{steps}}$ the number action steps, and $r_{\text{stage}}$ is either of three stages.
Stage~1 gives an increasing reward for proximity to the target position, aligning with the target angle and opening the grapple, and stage~2 gives an increasing reward for closing the grapple.
Stage~3 is activated once the grapple has closed around at least one log, with an increasing reward for lifting the grapple.
We believe that the use of a dense reward term is vital for learning appropriate grapple angles from image data, where the dense reward greatly increases the feedback as to what grapple angle the image data represents.
The third term in Eq.~(\ref{eq:reward}) is a penalty for excessive energy use, which is proportional to the sum of the power of the actuators.

\subsubsection{Curriculum}
\label{sec:curriculum}
Each episode of the RL task features a pile placed according to a function, with a \emph{difficulty parameter} $d \in [0, 1]$ determining the challenge level.
To speed up the simulations, we keep the vehicle in the same configuration and place the pile in relation to it.
For $d=0$, the pile is always placed just below the starting position of the grapple, while for $d=1$ it is placed with random rotation at challenging positions on either side of the vehicle at varying heights $z \in [-1/2, 1]$~m.
For intermediate difficulty levels, a linear interpolation of the two cases is used, allowing the challenge of the task to be smoothly adjusted.
Collisions between the vehicle and the crane/piles are disabled as piles can overlap with the vehicle, especially during the curriculum.

The curriculum consists of lessons where we adjust the difficulty parameter in increments of $0.1$.
20 evaluation episodes are conducted every 50,000 steps, and progress to the next lesson is determined by the mean accumulated reward of the past $10~\times 20$ evaluation episodes, compared to a threshold.
The threshold was empirically determined, set to $21$, to allow progress through the curriculum on a regular basis.
In addition to varying the target position, we also modify the criterion for target success.
The logs must be raised higher above the ground as the lessons become more challenging, from 0.25~m for $d=0$ to 1.1~m for $d=1$.

\subsubsection{RL Algorithm and Network}
We use the Stable-Baselines3~\cite{stable-baselines3} RL library with the \emph{model-free}, \emph{on-policy} algorithm PPO~\cite{schulman2017proximal}.
This can enable learning in complex environments, but tends to be sample inefficient.
Unlike model-based methods, it does not build an internal model of the environment but learns a mapping from states to actions to maximise the expected accumulated discounted reward.
After each policy update in PPO, new data must be acquired using the new policy.

The input data for our RL agent consists of 16 floating-point numbers and two channels of $64 \times 64$ images.
The images pass through a CNN, \emph{feature extractor} network and the resultant vector is concatenated with the other observations.
The concatenated input is then fed into two fully connected neural networks, one to predict the value function and the other to generate the action.

We train using 8 environments with a maximum episode length of 10~s, a simulation frequency of 60~Hz and a control frequency of 20~Hz.
A number of hyperparameters, such as batch size, learning rate, and network parameters, were varied to find the best-performing agents.
The best model was trained using a batch size of 1600, a learning rate of 0.00025, and a feature extractor CNN with (8, 8, 8) filters of sizes [8, 4, 3] and strides [4, 2, 1], and 64 output features.
The fully connected networks have two hidden layers of size (64, 64), with tanh activation functions.
A summary of the hyperparameters can be found in Table \ref{table:hyperparameters}.

\begin{table}[hth]
\begin{tabular}{l c l c }
  Hyperparameter  & value &   Hyperparameter  & value \\
\hline
n-envs & 8  & episode-length & 200 \\
batch-size & 1600 & learning-rate & 0.00025 \\
gamma & 0.99 & n-epochs & 4 \\
ent-coef & 0.0 & vf-coef & 0.5 \\
max-grad-norm & 0.5 & gae-lambda & 0.95 \\
clip-range & 0.2 & & \\
\end{tabular}

\caption{\label{table:hyperparameters}Hyperparameters used, for details see~\cite{stable_baselines_ppo}.}
\end{table}

\section{Results and Discussion}
We present results from training and evaluating an agent.
We analyse the importance of the observations and try to shed light on the inner workings of the agent.

\subsection{Training}
\begin{figure}[ht]
\centering
  \includegraphics[width=0.48\textwidth]{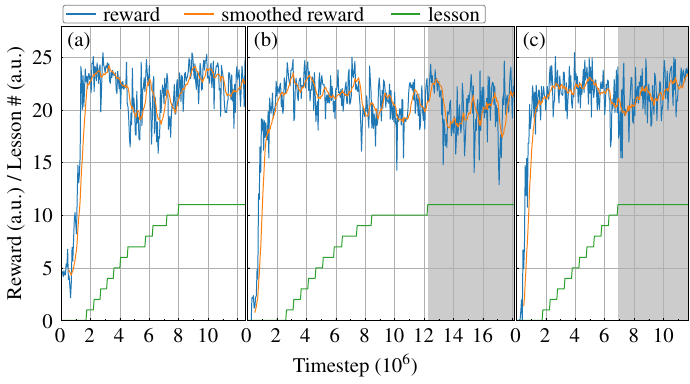}
  \caption{\label{fig:training_curves} Evaluation curves during training of the selected agent, showing reward, lesson number, and smoothed reward using a sliding window of size 10.
    (\textbf{a}) shows training with two logs and restricted radius range. (\textbf{b},\textbf{c}) shows training with 2--5 logs, with the grey regions highlighting the final lesson with non-simplified task. The lesson number maps to the difficulty parameter $d$, as described in Section~\ref{sec:curriculum}.}

\end{figure}

We train agents for up to $20$ million steps and extend the training of successful agents. Fig.~\ref{fig:training_curves} shows training curves for the selected agent, achieving the highest smoothed reward.
This was first trained on piles with two logs placed at a restricted radius range $r \in [4.5, 5.5]$.
Loading from the best-performing stage, training was resumed with piles of 2--5 logs and the full radius range.
After once more reloading from the best-performing stage and passing the curriculum, the best agent was selected.
Having passed through all the curriculum lessons, the agent reward and success rate fluctuate at a high level but without achieving consistent mastery.

\subsection{Evaluation}
\begin{figure}[ht]
\centering
  \includegraphics[width=0.9\columnwidth]{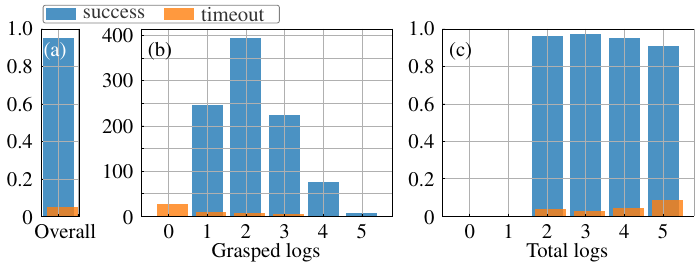}
  \caption{\label{fig:success} (\textbf{a}) shows an overall success of 95\%, (\textbf{b}) shows what number of logs is grasped, and (\textbf{c})~shows success given how many logs are in the pile.}
\end{figure}

\begin{figure}[ht]
\centering
  \includegraphics[width=0.48\textwidth]{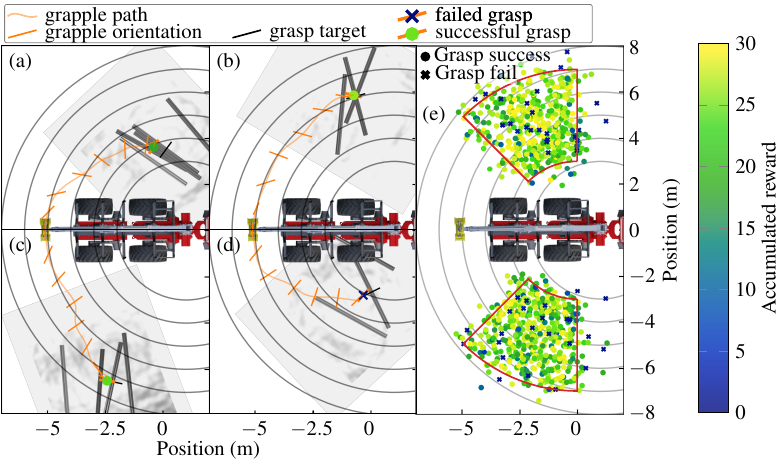}
  \caption{\label{fig:attempt_examples} (\textbf{a}--\textbf{d}) Example of four grasp attempts and (\textbf{e}) illustration of target locations.
The grapple path and orientations are shown in yellow, with suggested/actual grasp poses in black/thick yellow.
Target locations and grasps are coloured after accumulated reward and with failures illustrated by $\times$.
The red outline marks the region where piles are placed, with target locations outside of this due to offsets from pile centres.
}
\end{figure}

\begin{figure}[ht]
\centering
  \includegraphics[width=0.6\columnwidth]{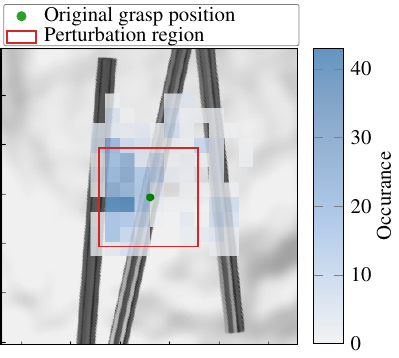}
  \caption{\label{fig:attempt_perturbations}
Heatmap showing the grasp position of the agent for 625 grasp attempts where the original target position is systematically perturbed, evenly within a $1\times 1$~m region, for the same pile as in Fig.~\ref{fig:attempt_examples}c.
  }
\end{figure}

\begin{figure}[ht]
\centering
  \includegraphics[width=0.9\columnwidth]{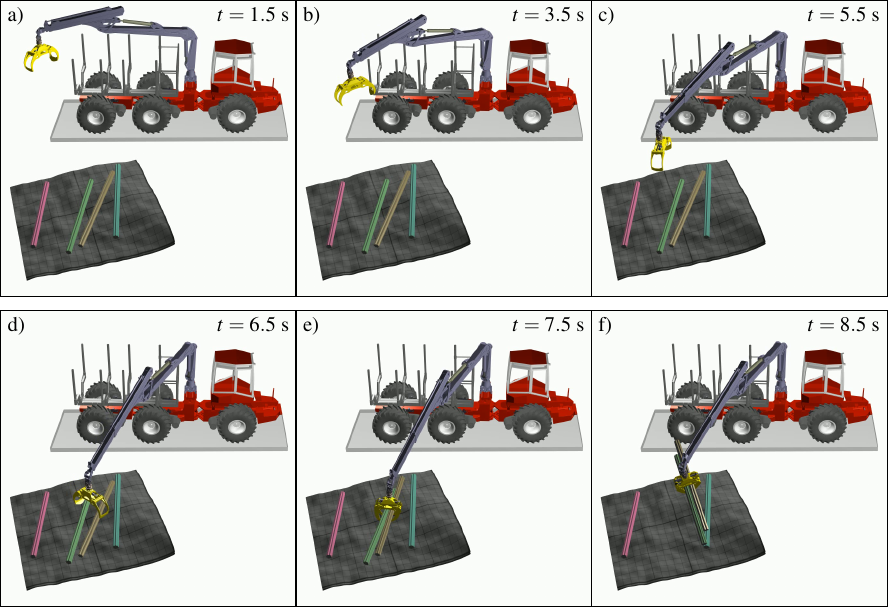}
  \caption{\label{fig:grasp_seq} (\textbf{a}--\textbf{f}) Image sequence of one example grasp, for the same pile as in Fig.~\ref{fig:attempt_examples}c.
  }
\end{figure}

To evaluate the agent's performance, we conducted 1000 grasp attempts on evaluation piles with 2--5 logs.
The test setup is similar to the training setup but with a set of piles that have not been used during training.
Success was defined as the agent grasping one or more logs and lifting them to an elevation gain of 1.1~m.
The overall success rate is $95$~\%, as shown in Fig.~\ref{fig:success}, and the most common yield is two grasped logs.
The success rate is the highest (97\%) for piles of three logs and the lowest (91\%) for piles of five.
Fig.~\ref{fig:attempt_examples}e shows the target grasp position for each attempt, coloured by the accumulated reward and with failed attempts shown as $\times$.
The agent has learned to pick logs over the entire area, and there is no apparent systematic pattern for the failures.
The design of the target distribution function sometimes results in logs close to/underneath the vehicle.
If collisions between the crane and the vehicle are enabled, there can be collisions for targets within 1.25~m from the wheels.
For low targets in the very back, the main boom can also collide with the load bunk due to an under-use of the telescope in the IK implementation.
As the grasp pose is set according to the position and orientation of one of the logs, this is most likely not the optimal pose for grasping multiple logs.
To increase the probability of picking multiple logs, the agent must thus learn to make deviations from the suggested grasp pose using the camera data.
Fig.~\ref{fig:attempt_examples} show details of four specific attempts, three successful and one failed.
One should be careful to draw conclusions from individual evaluations, as the agent and its interaction with the environment is complicated and can give rise to seemingly random behaviour.
Even so, Fig.~\ref{fig:attempt_examples}c seems to show a small deviation to better grasp two logs instead of one, and Fig.~\ref{fig:attempt_examples}b seems to show the grapple rotation adjusted to better grasp both logs.
To determine whether this is a coincidence or a learned strategy, we introduce systematic perturbations in the target grasp position and study the resultant spatial distribution of the actual grasp positions for the specific case of Fig.~\ref{fig:attempt_examples}c.
The resultant grasp positions are mainly drawn towards in between the leftmost logs or on the log to the right, as can be seen from the heatmap in Fig.~\ref{fig:attempt_perturbations}.
We conclude that the agent can utilise the camera data to make strategic deviations from a given target position.
Typical variations in grasp position compared to the target position for the 1000 evaluation grasps are in the range $\pm$0.5~m.
The agent is thus not more sensitive than so to the recommended grasp position.
The grasp sequence in a 3D view corresponding to the case Fig.~\ref{fig:attempt_examples}c can be seen in Fig.~\ref{fig:grasp_seq}.

\subsection{Observation Ablation Study}
To understand the importance of different observations for the agent, we conducted three types of ablation experiments.
Firstly, we trained the agent with and without particular observations to measure their importance.
Secondly, we added noise to some observations during evaluation and measured the resulting loss in performance.
Lastly, we added noise to observations in already recorded data and measured the resultant change in actions.

Retraining the agent without some observations or with additional observations is computationally expensive and time-consuming.
We, therefore, only perform these experiments selectively to verify some of the design decisions in setting up the agent, such as not providing the target angle as an observation to impose a greater dependency on the camera data.
We also use it to verify the agent's use of camera data.
There is a bias in favour of the baseline, in the sense that the hyperparameters and the curriculum have been set up to achieve success for the baseline, whereas this may not be optimal for the other cases.
Even so, where there are significant differences, this can still provide a decent indication of the importance of adding or removing an observation in this particular setup.
The results are measured in the total number of lessons passed, including repeatedly passing the final lesson, and can be seen in Table \ref{table:obs_ablation}.
It can be seen that adding the target angle as an observation is detrimental to learning.
This aligns with our idea that a dense dependence on camera data is crucial for learning to use it effectively.
Removing the depth camera seemed to be more challenging than removing the greyscale camera, indicating the importance of the depth camera in this stage of training.
Removing both cameras and instead relying on the target angle results in very poor training.
It is definitely possible to learn such a task, as shown in \textcite{andersson2021reinforcement} for a single log on flat ground, and the bad performance could be an example of the bias towards the baseline mentioned earlier.
However, it verifies the use and importance of the camera data in this particular setup and how the baseline agent learns features in the camera data not captured by the target angle alone.

\begin{figure}[h]
\centering
  \includegraphics[width=0.9\columnwidth]{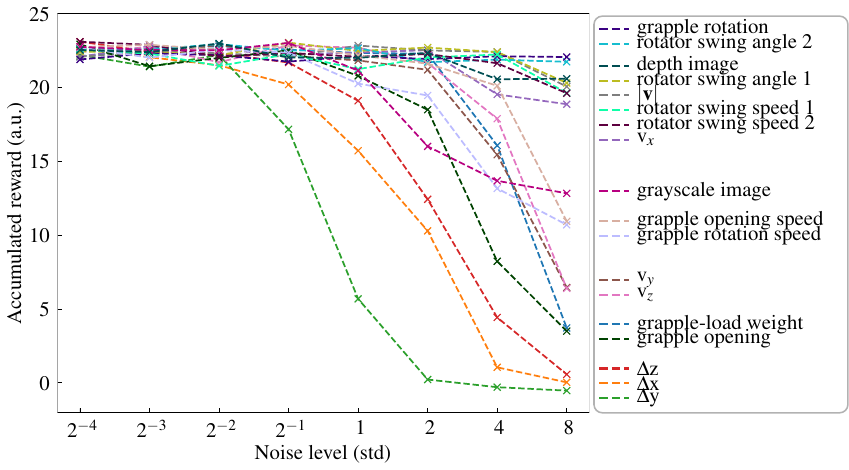}
  \caption{\label{fig:obs_noise_eval} Mean accumulated reward over 100 evaluations while adding different levels of noise to each observable in turn.}
\end{figure}

To gain further insight into the importance of different observations, we add different levels of noise to the observation signals during evaluations.
The idea behind this approach is that the reward will be sensitive to noise on important observations and insensitive to noise on less important or redundant observations.
There is no impartial level of noise that would enable a perfect comparison between different observations.
Still, we try to find fair levels based on the distribution of each observable's values.
We find the standard deviation $\sigma_i$ of each observation $o_i$ from an evaluation using 1000 grasp attempts and use it to scale the noise for each observation.
The added noise is drawn from a Gaussian distribution, and we consider noise at eight different levels, in the range $[2^{-4}, 2^{3}] \times \sigma_i$.
To determine the performance given an observation and level of noise, we perform 100 evaluations and measure the mean reward.
The results can be seen in Fig.~\ref{fig:obs_noise_eval}.
The relative position is clearly very important.
Other important observations are the grapple-load weight and the opening angle of the grapple.
In contrast, observations related to the swing angle and speed, as well as the rotation of the grapple, seem not to be important to the agent.
The latter shows how the grapple rotate action does not depend on the rotation angle, but rather the camera data as well as the rotation speed.
\begin{table}[ht]
\begin{tabular}{l l c}
\# &  Case & \multicolumn{1}{l}{lesson success (std)}  \\ 
\hline
\hline
0 &  baseline & 91.2~(39.7) \\ 
\hline
1 & $+$ target angle & 18.6~(10.2)  \\ 
2 & $+$ joint angles & 10.4~(2.0)  \\ 
3 & $-$ depth camera & 10.4~(6.0)  \\ 
4 & $-$ grayscale camera & 47.0~(30.9)  \\ 
5 & $-$ cameras, $+$target angle & 1.8~(3.6)  \\ 
\end{tabular}
\caption{Results of adding (+) or removing ($-$) observations on the total amount of lessons passed during 20~M steps of training.
The trainings were repeated 5 times, and the mean and standard deviation are displayed.}
\label{table:obs_ablation}
\end{table}

\begin{figure}[!htb]
\centering
  \includegraphics[width=0.48\textwidth]{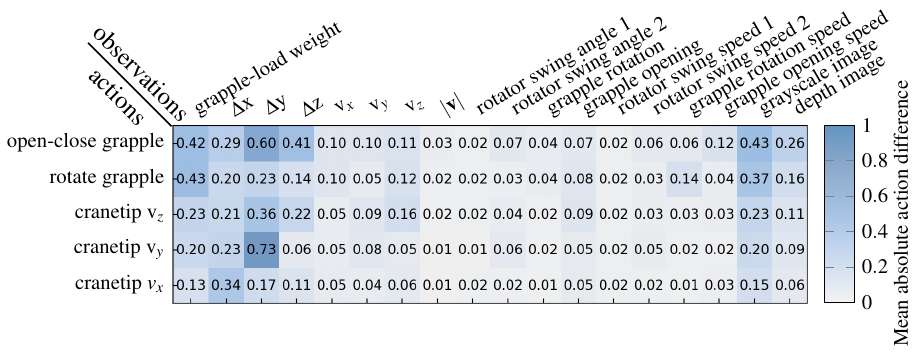}
  \caption{\label{fig:model_dependency} Mean absolute difference in action by adding noise to each observation in turn on recorded data from 1000 evaluations.}
\end{figure}

Finally, we use recorded data from 1000 evaluations and add noise to each observation in turn to observe how the actions of the agent change.
As well as highlighting important observations, this can tell when during the load cycle an observation is most important and for what actions.
This can give insights into the inner workings of the agent.
The noise is drawn from a Gaussian distribution with $\mu=0$ and $\sigma_i=0.2\left(\max(o_i) - \min(o_i)\right)$.
As can be seen from the results in Fig.~\ref{fig:model_dependency}, the importance of the relative position and the grapple-load weight is once again highlighted.
The relative position is obviously important for the crane-tip actions, and there is a clear importance for the open-close grapple action too.
Adding noise on the greyscale image channel has a larger effect on the actions than adding noise to the depth camera, which is consistent with the corresponding larger drop in reward seen in Fig.~\ref{fig:obs_noise_eval}.
However, the results in Table~\ref{table:obs_ablation} do suggest the depth camera is more important for passing the curriculum during training.
From Fig.~\ref{fig:model_dependency}, we see that the greyscale camera is important for positioning and aligning the grapple with the logs, as well as for timing the closing of the grapple, while the depth camera is mostly important for timing the closing of the grapple.
This is not unreasonable, considering that the greyscale camera shows greater contrast between the logs and the background, while the depth camera provides information about the distance to the logs.
It might be the case that the depth camera is more important for passing the curriculum, even if the agent in the latter stage of training has a greater dependence on the greyscale camera.

\section{Conclusions}
We conclude that using a virtual camera stream from 3D reconstructed data is a viable setup for multi-log grasping, with the agent using the camera data for grasping despite the underlying data not updating during the grasp as a real camera would.
The agent learns to pick logs with 95\% accuracy, using the camera when steering the crane tip, as well as rotating and closing the grapple.
The Cartesian control simplifies domain adaption for deploying the RL agent on a real machine.
Using a virtual camera allows for collecting visual information when the view is not occluded, combining data from different times or perspectives, and working with processed data to avoid real-time segmentation.
This enables solutions for problems related to segmentation, occlusion, season, weather, and light conditions in applications in unstructured forest environments.

The grasping agent has a modular design that is interoperable with any method for crane control that takes the crane-tip target velocity as input.
That includes existing methods for time-optimal trajectory planning and control \cite{ortiz2014increasing} and semi-autonomous shared control \cite{hansson2010semi} with the possibility of introducing geo-fences around the machine and other known objects.
This interoperability is important to ensure the safety and productivity of the automated system, e.g., through human monitoring of the planned motion with the possibility of intervening by manually adjusting the speed and direction of the crane-tip motion.
The implication is that automatic loading can be introduced as an assistive system well before the system is mature for autonomous control.

The observation ablation/augmentation study gave insights into the inner workings of the agent, showing how a dense dependence on camera data is important for the agent to utilise vision and how the agent uses features of camera data that are not captured by the target angle alone.
The observation noise study revealed the importance of each observation.
It indicated that the grapple-load weight is a vital observation and that the greyscale camera is more important than the depth camera for the trained agent.
Additionally, the study showed that the grapple rotate action is controlled by camera data and rotation speed, but does not involve the rotation angle itself.

Possible future work involves improvements in RL methods and training to achieve master-level performance, the inclusion of models for optimal grasp poses, the inclusion of log diversity in terms of size and shape, and the transfer of the learned skills to a real machine.
Transfer tests of the learned skills to a real machine will involve integrating with a log segmentation algorithm, such as the one described in \cite{fortin2022instance}, and interfacing with a crane control system that takes the crane-tip velocity as an input.
In addition to RGB-D sensing, the test system will need to be equipped with sensors for the grapple's orientation and opening and an estimator for the load weight.

\section{Acknowledgements}
This work was supported by
Mistra Digital Forest, Algoryx Simulation AB, Cranab AB, and eXtractor AB.
The simulations were performed on resources provided by the Swedish National Infrastructure for Computing at High Performance Computing Center North (HPC2N).

\printbibliography

\end{document}